\documentclass[lettersize,journal]{IEEEtran}
\usepackage{amsmath,amsfonts}
\usepackage{algorithm}
\usepackage{algorithmicx}
\usepackage{algpseudocode}

\usepackage{array}
\usepackage[caption=false,font=normalsize,labelfont=sf,textfont=sf]{subfig}
\usepackage{textcomp}
\usepackage{stfloats}
\usepackage{url}
\usepackage{verbatim}
\usepackage{graphicx}
\usepackage{cite}
\usepackage{color}
\usepackage{graphicx} 
\usepackage{url}

\newcommand{\orcid}[1]{
\href{https://orcid.org/#1}{\includegraphics[width=10pt]{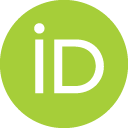}}
}

\begin{document}

\title{Exploring Adversarial Examples and Adversarial Robustness of Convolutional Neural Networks by Mutual Information}
% \author{}
\author{Jiebao Zhang, Wenhua Qian,  Rencan Nie, Jinde Cao, \IEEEmembership{Fellow,~IEEE}, and Dan Xu
% \author{Wenhua Qian,~\IEEEmembership{Staff,~IEEE,}
        % <-this % stops a space
\thanks{This work was supported by the Research Foundation of Yunnan Province No.202002AD08001, 202001BB050043,
2019FA044, National Natural Science Foundation of China under Grants No.62162065, Provincial Foundation for Leaders of Disciplines in Science and Technology No.2019HB121, in part by the Postgraduate Research and Innovation Foundation of Yunnan University (No.2021Y281, No.2021Z078), and in part by the Postgraduate Practice and Innovation Foundation of Yunnan University (No.2021Y179, No.2021Y171). \textit{(Corresponding author: Wenhua Qian.)}}% <-this % stops a space
\thanks{Jiebao Zhang, Wenhua Qian, Rencan Nie, and Dan Xu are with the School of Information Science and Engineering, Yunnan University, Kunming 650500, China (e-mail: zhangjiebao2014@mail.ynu.edu.cn; whqian@ynu.edu.cn).}
\thanks{Wenhua Qian and Rencan Nie are also with the School of Automation, Southeast University, Nanjing 210096, China.}
\thanks{Jinde Cao is also with the Department of Mathematics, Southeast University,
Nanjing 210096, China.}
% \thanks{Manuscript received April 19, 2021; revised August 16, 2021.}
}

% The paper headers
\markboth{Journal of \LaTeX\ Class Files,~Vol.~14, No.~8, August~2021}%
{Shell \MakeLowercase{\textit{et al.}}: A Sample Article Using IEEEtran.cls for IEEE Journals}

% \IEEEpubid{0000--0000/00\$00.00~\copyright~2021 IEEE}
% Remember, if you use this you must call \IEEEpubidadjcol in the second
% column for its text to clear the IEEEpubid mark.

\maketitle

\begin{abstract}
A counter-intuitive property of convolutional neural networks (CNNs) is their inherent susceptibility to adversarial examples, which severely hinders the application of CNNs in security-critical fields.
Adversarial examples are similar to original examples but contain malicious perturbations.
Adversarial training is a simple and effective defense method to improve the robustness of CNNs to adversarial examples.
The mechanisms behind adversarial examples and adversarial training are worth exploring.
Therefore, this work investigates similarities and differences between normally trained  CNNs (NT-CNNs) and adversarially trained CNNs (AT-CNNs) in information extraction from the mutual information perspective.
We show that
1) whether  NT-CNNs or AT-CNNs, for original and adversarial examples, the trends towards mutual information are almost similar throughout training; 
2) compared with normal training, adversarial training is more difficult and the amount of information that AT-CNNs extract from the input is less;
3) the CNNs trained with different methods have different preferences for certain types of information; NT-CNNs tend to extract texture-based information from the input, while AT-CNNs prefer to shape-based information.
The reason why adversarial examples mislead CNNs may be that they contain more texture-based information about other classes.
Furthermore, we also analyze the mutual information estimators used in this work and find that they outline the geometric properties of the middle layer's output\footnote{Code: \url{https://github.com/wowotou1998/exploring-adv-by-mutual-info}
}.
% \footnote{1111111111111}
% 本工作探究了在不同训练策略下, 样本与模型内部层输出的互信息变化情况,
% we explore the mutual information changes between middle layer output and inputs, and between middle layer output and labels in.
% 这里应该说对抗样本也包含有很多的信息,
% 这里可以加一点说明对抗样本会让神经网络分类失效的原因,对抗训练更鲁棒的原因/或者做一下总结

\end{abstract}


\section{Introduction}
% 开头介绍对抗样本,对抗训练
\IEEEPARstart{C}{onvolutional} Neural Networks (CNNs) have achieved surpassing performance on many tasks \cite{deep_learning_nature}, \textit{e.g.}, image classification \cite{image_classification_survey}, object detection \cite{object_detection_survey}, style transfer \cite{style_transfer_survey}, image captioning \cite{image_captioning_survey}, and other fields. However, researchers have found that adding small malicious perturbations crafted by adversaries to the original examples can drastically degrade the performance of CNNs \cite{evasion_attack,L-BFGS_attack}. This makes CNNs unreliable for security-sensitive tasks. There are many methods to improve the resistance of CNNs against adversarial examples. Adversarial training (AT), a training paradigm, is one of the most effective defense methods \cite{FGSM, PGD, Random-FGSM}.
Compared to the objective of previous normal training that minimizes the empirical risk of the original dataset, adversarial training pursues the empirical risk minimization over the modified dataset containing malicious perturbations.
% 这里引入对对抗样本和对抗训练为什么起作用的一些探究工作
\par
Where the vulnerability of CNNs to adversarial examples stems from and why adversarial training can improve the robustness of CNNs are worth exploring. At first, Szegedy \textit{et al.} argues that the vulnerability of CNNs to adversarial examples is caused by the local linearity of the neural networks \cite{FGSM}. 
Recent works claim that the vulnerability could be attributed to the presence of non-robust features \cite{adversarial_examples_are_features}. 
The robustness of adversarially trained CNNs is a consequence of the fact that models learn the representations aligned better with human perception, namely the shape-based representation (\textit{e.g.}, shapes and contours in images) \cite{robustness_may_be_odds_with_accuracy}.
Many experimental results explicitly or implicitly support this view.
The BagNet \cite{BagNets} is a simple variant of the ResNet \cite{ResNet} with limiting the receptive field size of the topmost convolutional layer.
Its performance and uncomplicated architecture suggest that current network architectures base their decisions on relatively weak and local statistical regularities of inputs \cite{BagNets}. For example, the number of wheels in a image determines whether it is a bicycle or a car.
% \cite{BagNets} argues that it is caused by the local patterns learned by the neural network itself. 
Furthermore, Geirhos \textit{et al.} and Hermann \textit{et al.} demonstrate that ImageNet-trained CNNs are inclined to make decisions by the texture-based representation rather than shape-based representation \cite{texture_bias_in_CNN_2019,texture_bias_in_CNN_2020}. 
Echoing the literature mentioned above, Zhang \textit{et al.} also show that adversarial training can change the texture-based representation bias of CNNs and force CNNs to appreciate more the shape-based information \cite{interpreting_AT-CNN}. 

% 这里引入信息瓶颈理论, 讲述从信息论的视角来探究神经网络
\par
Moreover, some researchers have tried to explain the working mechanism of deep neural networks (DNNs), \textit{e.g.}, information bottleneck (IB) principle\cite{information_bottleneck_1}. In IB principle, given two random variables $X$ (\textit{e.g.}, images) and $Y$ (\textit{e.g.}, labels), $Y$ implicitly determining the relevant and irrelevant features in $X$, an optimal representation $\hat{X}$ of $X$ can capture the relevant features about predicting $Y$ \cite{information_bottleneck_1}. 
The IB principle provides a new information-theoretic optimization criteria for optimal DNN representations \cite{information_bottleneck_2}.
The trends towards the mutual information in DNNs estimated by the binning method appear to report the plausibility of the IB principle \cite{information_bottleneck_3}.
Meanwhile, inspired by the IB method, deep variational information bottleneck (VIB) \cite{variational_information_bottleneck} and non-linear 
information bottleneck (NIB) \cite{nonlinear_information_bottleneck} objectives are proposed. 
Compared with deterministic models with various forms of regularization, the models trained with the VIB or NIB objective have better performance concerning generalization and resistance to adversarial attacks \cite{variational_information_bottleneck}.
However, there are some arguments about the applicability and conclusions derived from the IB principle \cite{investigate_information_bottleneck_functional,information_plane_review}. 
Saxe \textit{et al.} demonstrate that the information plane trajectory is predominantly a function of the nonlinear activation and the double-sided saturating nonlinear function (\textit{e.g.}, tanh) yields the compression phase \cite{opposing_view_to_information_bottleneck}.
Goldfeld \textit{et al.} claim that the compression phenomenon throughout training is driven by progressive geometric clustering of the middle representation \cite{estimate_MI_flow_in_DNNs}.
\par
\begin{figure*}[!t]
	\centering
	\includegraphics[width= \textwidth]{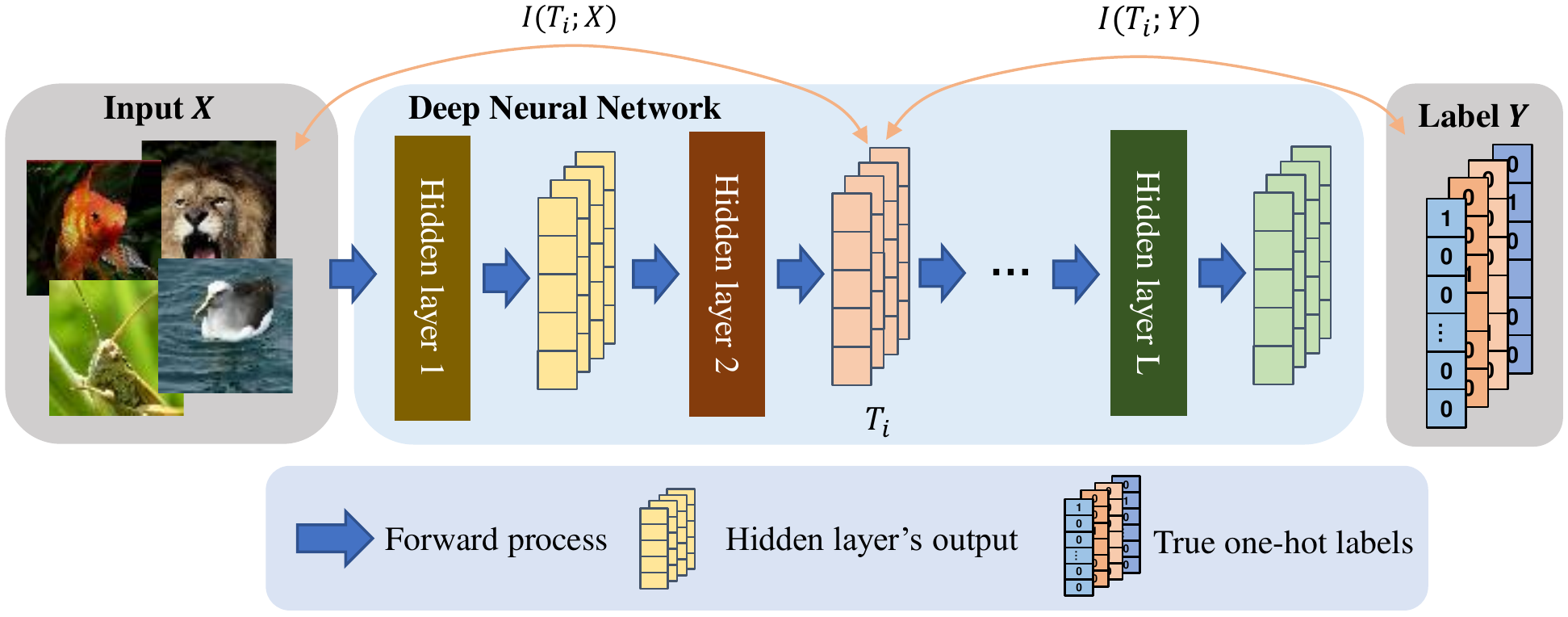}
	\caption{An overview of the entire mutual information estimation. The mutual information $I(T_i; X)$ between the $i$-th middle layer's output $T_i$ and the input $X$. 
	The mutual information $I(T_i; Y)$ between the $i$-th middle layer's output $T_i$ and the label $Y$.}
	\label{fig: work_flow}
\end{figure*}

Here we put these debates about the IB principle aside. 
In this work, we get inspiration from the way that the mutual information is used to analyze the relevant information between layers in the IB principle. We study the similarities and differences between normally training CNNs (NT-CNNs) and adversarially training CNNs (AT-CNNs) from the mutual information perspective, 
The relevant information is quantified during the training process between the middle layer's output and the example, and between the middle layer's output and the label.
Specifically, for a label variable $Y$, an input variable $X$, and a corresponding $i$-th layer's output $T_i$ of a model, we can estimate the mutual information $I(T_i; X)$ and $I(T_i; Y)$ on each training epoch, as shown in Fig. \ref{fig: work_flow}.
Whether NT-CNNs or AT-CNNs, we feed original and adversarial examples into models and explore the trends towards $I(T_i;  X)$ and $I(T_i;  Y)$ throughout training.
We have some observations after analyzing the experimental results. First, whether NT-CNNs or AT-CNNs, for original and adversarial examples, the trends towards mutual information are almost similar throughout training; second, compared with normal training, adversarial training is more difficult and the amount of information that AT-CNNs extract from the input is less; and third, the CNNs trained with different methods have different preferences for certain types of information; NT-CNNs tend to extract texture-based information from the input, while AT-CNNs prefer to shape-based information.
\par
The vulnerability of NT-CNNs to adversarial examples may derive from the extraction preferences about some types of information. 
% That is which type of information models tend to extract from inputs.
The reason why adversarial examples easily mislead NT-CNNs may be that they contain more texture-based information about other categories, while NT-CNNs prefer to extract texture-based one.
Conversely, AT-CNNs' predictions predominately depend on shape-based information.
The fact that small perturbations in adversarial examples do not easily change the shape-based information may bring about robustness in AT-CNNs.
Furthermore, we find that the mutual information estimators outline some properties of the middle layer's output from the geometric perspective. The tighter the middle layer's output cluster, the smaller the mutual information $I (T_i; X)$.
When the middle layer's output distribution is determined, the tighter the middle layer's output belongs to one class cluster, the larger the mutual information $I (T_i; Y)$.
\par
In summary, this article has made the following contributions.
\begin{itemize}
    \item{We explore the trends towards $I(T_i;  X)$ and $I(T_i;  Y)$ on NT-CNNs and AT-CNNs during the training phase. The empirical results show whether the input is the original or adversarial example, the mutual information trend of AT-CNN nearly coincides with the one of NT-CNN.}
    \item{We analyze the information curves about $I(T_i;  X)$ and $I(T_i;  Y)$ on NT-CNNs and AT-CNNs when the original and adversarial input suffers information distortion. The empirical results demonstrate that CNNs trained with different training paradigms indeed arise the information extraction bias.}
    \item{We uncover that the mutual information estimators used in this work practically outline some geometric properties of the middle layer's output after analyzing key parts of algorithms.}
\end{itemize}
\par
The rest of this paper is organized as follows. 
Section \ref{sect: related_work} reviews some literature on adversarial attacks and adversarial training methods. 
Section \ref{sect: methods} elaborates on the mutual information estimators. 
Section \ref{sect: experiments} illustrates extensive experimental results based on various models and datasets.
Section \ref{sect: discussion} analyzes the mutual information estimators used in this work and concludes some viewpoints from the geometric perspective.
Finally, conclusions are given in Section \ref{sect: conclusion}.
% In this work, we apply the information-theoretic
% IB principle can illustrate and guide the training process of CNNs.The experimental results in \cite{information_bottleneck_3} 
% 在本文中, 我们把标签看成随机变量Y,输入看成随机变量X, 与此对应的各中间层的输出看成随机变量T_i, 使用互信息估计器来估计中间层T_i和随机变量x,

\section{Related Work}
\label{sect: related_work}
\subsection{Adversarial Attacks}
In the inference phase, when examples contain human-imperceptible and malicious perturbations, the classification performance of DNNs shows a sharp decrease \cite{evasion_attack,L-BFGS_attack}.
This inherent weakness of DNNs arouses the interest of researchers. 
Many adversarial attack methods have been proposed to seek perturbations according to the exclusive attributes of DNNs and optimization techniques.
Adversarial attacks can be divided into two categories: white-box and black-box attacks. White-box attacks can be further divided into optimization-based attacks \cite{L-BFGS_attack,C&W}, single-step attacks \cite{FGSM,Random-FGSM}, and iterative attacks \cite{I-FGSM,DeepFool,PGD,MI-FGSM}. 
\par
Optimization-based attacks formulate finding an optimal perturbation as a box constraint optimization problem.
Experiments show that optimization-based attacks can achieve excellent attack performance.
The L-BFGS attack uses the second-order Newton method to solve this problem \cite{L-BFGS_attack}. 
Compared with the L-BFGS attack, the C\&W attack uses variable substitution to bypass the box constraint, replace the objective with a more powerful one, and uses the Adam optimizer \cite{Adam} to solve the optimization problem \cite{C&W}, the expression as follows:
\begin{equation}
\label{eq: cw_attack}
\begin{gathered}
\underset{\delta}{\operatorname{minimize}} \quad\| \delta \|_{p}+c \cdot f(x+ \delta) \\
\text {subject to} \quad x+ \delta \in[0,1]^{n} \\
f ( \cdot )=
\max 
\left(
\max 
\left\{
Z( \cdot )_{i, i \neq t}
\right\}-
Z( \cdot )_{t},-\kappa
\right) \\
\delta =\frac{1}{2}(\tanh (w)+1)-x,
\end{gathered}
\end{equation}
where $\delta$ denotes the perturbation vector, $x$ is original input, $Z(\cdot)$ represent the target DNN, $t$ is the target label index and $\kappa \geq 0$ is a tuning parameter for attack transferability.
\par
Single-step attacks are straightforward and effective to avoid the high computation costs caused by optimization attacks. Since models are assumed to be locally linear in single-step attacks, the perturbation is directly added along the gradient direction of the model to the original examples.
\par
Iterative attacks achieve the trade-off between computation and attack performance. Iterative attacks add perturbations several times. After each gradient is calculated, perturbations are added. The MI-FGSM incorporates the momentum term to stabilize the perturbation direction, which promotes the attack performance and the transferability of adversarial examples \cite{MI-FGSM}. The PGD attack does not directly add perturbations to the original example but selects a substitute example in the example's vicinity for subsequent operations \cite{PGD}, formulated as \begin{equation}
\label{eq: PGD}
x^{k+1}=\Pi_{x+\mathcal{S}}\left(x^{k}+\alpha \operatorname{sgn}\left(\nabla_{x} L(\theta, x, y)\right)\right),
\end{equation}
where $y$ is the label,  $\theta$ denotes the model parameters, $L$ is the loss function of DNNs, $\alpha$ is the step size in attacks, $\mathcal{S}$ denotes the set of allowed perturbations, $\Pi$ is the projection operation, and $x^k$ represents the example endures $k$ perturbations. 
The DeepFool approximates general non-linear classifiers to affine multiclass classifiers to construct the misclassification boundary and find the minimum perturbations to move examples to the nearest misclassification boundary \cite{DeepFool}. The ability of white-box attacks paves the way for adversarial training.
\par
Transfer-based attacks \cite{transferability_in_ML,transferability_in_large_models_datasets} take various forms.
Since researchers have found that the attack ability of adversarial examples is transferable \cite{L-BFGS_attack,FGSM,universal_attack,transferability_in_ML,transferability_in_large_models_datasets}. Namely, adversarial examples crafted for a certain model may also have the attack effect on other models,
Using this transferability, adversaries can use substitute models to generate adversarial examples without the knowledge of the specific structure and parameters of the target model to be attacked. So, the adversarial examples generated by the above white box attack are all transferable.
\par
Black-box attacks mean that the adversaries barely have the information of the structures and parameters of the target model, but can query the output of models to examples that attackers feed into models. 
The OPA attack \cite{OPA} is a black-box attack based on the differential evolution algorithm. When only the confidence score of classifiers to the images is known, this attack method only needs to change a few pixels in images to achieve a strong attack effect. 
The ZOO-attack \cite{ZOO_attack} is also a black-box attack, based on zeroth-order optimization. 
It approximates the gradient by querying the model's confidence score to the input and imitates the optimization-based attack. 
The Boundary attack \cite{Boundary_attack}, which solely relies on the final model decision, resorts to the random walk and rejection sampling techniques. The performance is competitive with that of gradient-based attacks.

\subsection{Adversarial Training}
Goodfellow \textit{et al.} initially proposed adversarial training \cite{FGSM}, and PGD-AT \cite{PGD} has become the mainstream method. The PGD-AT formulates how to find the parameters of a robust model as a saddle point problem:
\begin{equation}
    \label{eq: PGD-AT}
    \begin{gathered}
    \min _{\theta} \left( \mathbb{E}_{(x, y) \sim \mathcal{D}}\left[\max _{\delta \in \mathcal{S}} L(\theta, x+\delta, y)\right] \right),
\end{gathered}
\end{equation}
where $(x,y)$ denotes the image-label pair that is sampled from dataset $\mathcal{D}$.
The solution to the inner maximization problem is to find the adversarial example with a high loss. The outer minimization is to update the model parameters to remove the effect of adversarial attacks.
\par
The FreeAT \cite{FreeAT} optimizes the training speed. At each step of generating adversarial examples, it reuses the gradient of the model on immature adversarial examples to update the model parameters, shorten the epochs of adversarial training and reduce the computation.
The YOPO \cite{YOPO} regards a DNN as a dynamic system, considers that the layers are uncoupled, and accelerates the training speed by limiting the number of forward and backward propagations.
Compared with the improvement of training speed by the FreeAT and YOPO, the FreeLB \cite{FreeLB} pays more attention to the optimization effect of each min-max process. It is similar to PGD-AT with slight differences. The FreeLB first obtains the average of the gradients of the model to immature adversarial examples in the $k$ iterations, and then the parameters of the model are updated by the average of the gradients.

\section{Estimate the mutual information in deterministic CNNs}
\label{sect: methods}
\subsection{Preliminaries}
\label{sect: preliminaries}
A DNN can be regarded as a complicated function nested by some simple functions (\textit{e.g.}, fully-connected layers, convolutional layers) $f_1,f_2, \cdots, f_L$, where $f_i$ accept the output of $f_{i-1}$, then the output of $f_i$ as the input of $f_{i+1}$.
For an input variable $X$ (\textit{i.e.}, the image), a label variable $Y$, the corresponding $i$-th middle layer's output is $T_i$, calculated by
\begin{equation}
	T_i=f_i (f_{i-1}(\cdots f_1(X) \cdots )), i=1,2,\cdots,L.
	\label{eq: middle_layer_output}
\end{equation}
Especially, the input of $f_{1}$ is $X$, and the output of $f_{L}$ represents the final output of a DNN. Obviously, for deterministic DNNs, $T_i$ is a function of $X$.
\par
% \textcolor{red}{there should discuss the discrete entropy and differential entropy. and we need to explain the conditional entropy.}
For two random variables $X$ and $Y$, the joint probability mass function $p(x, y)$ and marginal probability mass functions $p(x)$ and $p(y)$, the entropy $H(X)$  is a measure of the uncertainty of the random variable $X$.
Notably, for brevity, we use $H$ to denote both the discrete entropy and differential entropy in this work.
When $X$ is known, the remaining uncertainty of $Y$ is quantified by conditional entropy $H(Y|X)$.
The relative entropy is a measure of the distance between two distributions. 
The mutual information $I(X;Y)$ is the relative entropy between the joint distribution $p(x, y)$ and the product distribution $p(x)p(y)$ \cite{book_elements_of_information_theory}:
\begin{equation}
\begin{aligned}
I(X ; Y) &=\sum_{x \in \mathcal{X}} \sum_{y \in \mathcal{Y}} p(x, y) \log \frac{p(x, y)}{p(x) p(y)} \\
&=D(p(x, y) \| p(x) p(y)) \\
&=E_{p(x, y)} \log \frac{p(X, Y)}{p(X) p(Y)}.
\end{aligned}
\end{equation}

From another perspective, the mutual information $I(X; Y)$ is the reduction in the uncertainty of $X$ (or $Y$) due to the knowledge of $Y$ (or $X$), formalized as
\begin{equation}
    \begin{aligned}
    &I(X;Y) \\
    &= H(X)-H(X \mid Y) \\ 
    &=-\sum_{x} p(x) \log p(x) + \sum_{x} p(x) \sum_{y} p(y \mid x) \log p(y \mid x) \\
    &=- \int p(x) \log p(x) dx + \int p(x) dx \int p(y \mid x) \log p(y \mid x) dy.
\end{aligned}
\end{equation}

\subsection{Non-parametric Mutual Information Estimators}
\label{sect: mutual_info_estimators}
\begin{figure}[]
	\centering
	\includegraphics[width= \linewidth]{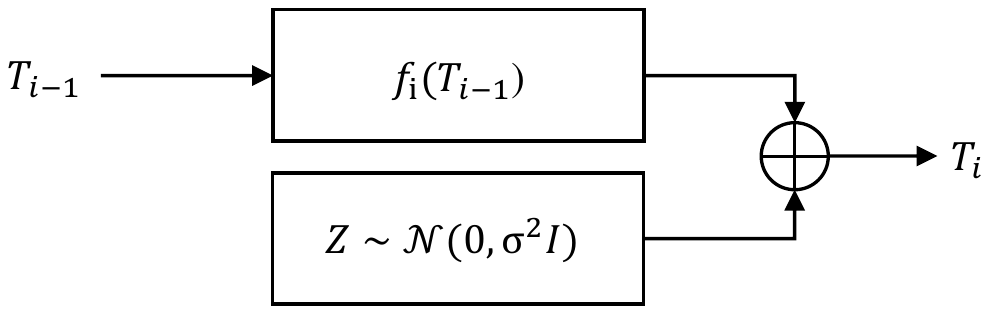}
    % 	Z 服从独立同分布的对角矩阵
	\caption{Illustration of the noise injection. $\sigma^2$ denotes the variance, $I$ is the identity matrix and $\oplus$ is the addition operation in an element-wise manner.}
	\label{fig: noise_injection_plot}
\end{figure}
\begin{figure}[]
	\centering
	\includegraphics[width= \linewidth]{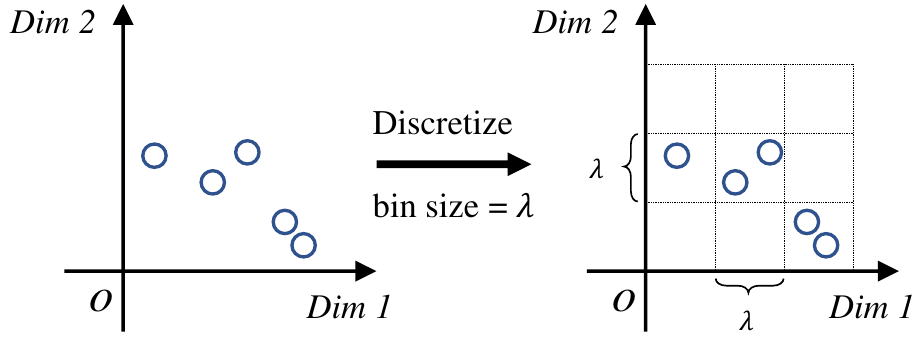}
	\caption{Illustration of the binning method. $\lambda$ denotes the bin size.
	The blue circles represent the sampling result of the 2-dimensional random variable $X$ according to the probability $P(X)$, and each cell represents the 2-dimensional bin. By calculating the number of data points in each bin, we obtain the empirical distribution and calculate the discrete entropy $H(X)$  through the empirical distribution. $H(X)=-\left(\frac{1}{5} \log \frac{1}{5}+\frac{2}{5} \log \frac{2}{5}+\frac{2}{5} \log \frac{2}{5}\right)$.}
	\label{fig: binning_method_plot}
\end{figure}
% For simplicity and ease of understanding, we only use several data points to represent mini-batch data. 

% 这一部分主要介绍在确定性网络中估计互信息的问题,
As mentioned in Section \ref{sect: preliminaries}, in this work, we have to estimate the mutual information in deterministic DNNs. For discrete variables, $I(T_i;X)$ between the $i$-th hidden layer's output $T_i$ and input $X$, is actually equal to $H(T_i)$, and for continuous variables, $I(T_i;X)$ is infinity. The details are shown in Appendix-\ref{appendix: A} and Appendix-\ref{appendix: B}. 
% 补充为什么在确定性神经网络中估计互信息是无穷大的,
% 还是得说清楚估计的是第几层的互信息, T_i
\par
\textbf{The kernel density estimation (KDE)}. The KDE estimators are proposed to estimate the lower and upper bounds of the entropy of mixture distributions (\textit{e.g.}, Gaussian mixture distribution) \cite{estimating_mixture_entropy}. 
In \cite{estimate_MI_flow_in_DNNs, nonlinear_information_bottleneck,opposing_view_to_information_bottleneck}, researchers transform deterministic DNNs into stochastic ones by artificially injecting Gaussian noise into the middle layer's output to avoid the infinite value when estimating the mutual information.
Concretely, in deterministic DNNs, the middle layer's output $T_i=f_i(T_{i-1})$, but in stochastic DNNs, $T_i=f_i(T_{i-1}) + Z$, $Z$ is Gaussian noise, and $ Z \sim \mathcal{N}(0,\sigma ^2 I)$, as shown in Fig. \ref{fig: noise_injection_plot}. 
Following the setting \cite{opposing_view_to_information_bottleneck}, in this work the noise is added solely when estimating mutual information and is not present during the training or testing phase.
The middle layer's output of a mini-batch can be regarded as a multivariate Gaussian mixture distribution in estimation, then we have
\begin{equation}
    \begin{aligned}
        I(T_i;X) &= H(T_i)-H(T_i \mid X) \\
                &= H(T_i)-H(Z).
    \end{aligned}
\end{equation}
When $H(Z)$ is directly computed and the upper and lower bounds of the entropy of $H(T_i)$ are estimated by KDE methods, the upper and lower bounds of mutual information $I(T_i; X)$ will be known.
The formula is as follows:
\begin{equation}
\label{eq: itx_by_kde_method}
\begin{aligned}
&\hat{I}(T_i;X) \\
&= \hat{H}(T_i) - \hat{H}(T_i \mid X)  \\
&= -\frac{1}{N} 
\sum_{j} 
\log \frac{1}{N} 
\sum_{k} 
\exp \left(-\frac{1}{2} \frac{\left\|t^{(j)}_{i}-t^{(k)}_{i}\right\|_{2}^{2}}{\gamma \cdot \sigma^{2}}\right), \\
\end{aligned}
\end{equation}
where $N$ denotes the number of samples in a mini-batch, $t^{(j)}_{i}$ is the output of $i$-th hidden layer for sample $j$, and $\gamma=4$ for computing the lower bound ($\gamma=1$ for computing the upper bound).
Furthermore, the lower and upper bounds of $I(T_i; Y)$ are as follows:
\begin{equation}
\label{eq: ity_by_kde_method}
\begin{aligned}
& \hat{I}(T_i; Y) \\
&= \hat{H}(T_i) - \hat{H}(T_i \mid Y)\\
&
=-\frac{1}{N} 
\sum_{j} 
\log \frac{1}{N} 
\sum_{k} 
\exp \left(-\frac{1}{2} \frac{\left\|t^{(j)}_{i}-t^{(k)}_{i}\right\|_{2}^{2}}{\gamma \cdot \sigma^{2}}\right) -\sum_{m} \frac{N_{m}}{N}
\\
&
\left[
-\frac{1}{N_{m}}
\sum_{j \atop y^{(j)}=m} 
\log \frac{1}{N_{m}} 
\sum_{k \atop y^{(k)}=m}
\exp 
\left(
-\frac{1}{2} 
\frac{\left\|t^{(j)}_{i}-t^{(k)}_{i}\right\|_{2}^{2}}{\gamma \cdot \sigma^{2}}
\right)
\right],
\end{aligned}
\end{equation}
where $N_m$ is the number of samples with class label $m$ and $y^{(j)}$ is the corresponding label for $x^{(j)}$.
More details for estimating $I(T_i;X)$ and $I(T_i;Y)$ by KDE methods are shown in Algorithm \ref{alg: KDE_methods}.

\textbf{The binning method}. The binning method can be adopted to discretize continuous values (\textit{e.g.},  the middle layer's output), and indirectly avoid the infinite differential entropy of continuous random variables. 
In \cite{information_bottleneck_3}, the layers' arctan output activations are binned into 30 equal intervals between -1 and 1. 
After counting the activations in each interval, the empirical joint distribution and marginal distributions can be calculated for computing the mutual information, a simple illustration shown in Fig. \ref{fig: binning_method_plot}.
As the bin size becomes smaller, the empirical distribution from sampling results of continuous variables will be more accurate but closer to the uniform distribution \cite{estimate_MI_flow_in_DNNs}.
More details for estimating $I(T_i; X)$ and $I(T_i;Y)$ by the binning method are shown in Algorithm \ref{alg: binning_method}.

\begin{algorithm}
	\caption{Estimate $I(T_i;X)$ and $I(T_i;Y)$ by KDE methods}
	\label{alg: KDE_methods}
	\begin{algorithmic}[1]
	    \Require{$N$ $i$-th middle layer's output $T_i$ in a mini-batch, $t^{(1)}_{i}, t^{(2)}_{i}, \cdots, t^{(N)}_{i}$, $N$ corresponding labels $y^{(1)}, y^{(2)}, \cdots, y^{(N)}$, the bound selector $\gamma$.}
	    \Ensure{The estimated mutual information, $\hat{I}(T_i;X)$ and $\hat{I}(T_i;Y)$.}
	    
		\State{Calculate $\hat{I}(T_i;X)$ by \eqref{eq: itx_by_kde_method};}
		\State{Calculate $\hat{I}(T_i;Y)$ by \eqref{eq: ity_by_kde_method};} \\
		\Return {$\hat{I}(T_i;X)$ and $\hat{I}(T_i;Y)$.}
	\end{algorithmic}
\end{algorithm}

\begin{algorithm}[h]
        \caption{Calculate $I(T_i;X)$ and $I(T_i;Y)$ by the binning method}
        \label{alg: binning_method}
        \begin{algorithmic}[1] %每行显示行号
            \Require {$N$  $i$-th middle layer's output $T_i$ in a mini-batch, $t^{(1)}_{i}, t^{(2)}_{i}, \cdots, t^{(N)}_{i}$, and  corresponding labels $y^{(1)}, y^{(2)}, \cdots, y^{(N)}$, total number of classes $M$, the binning interval size $\lambda$.}
            \Ensure {The estimated mutual information, $\hat{I}(T_i;X)$ and $\hat{I}(T_i;Y)$.}
            \State{Initialization;}
            \State{$T_i=\left[t^{(1)}_{i}, t^{(2)}_{i}, \cdots, t^{(N)}_{i}\right]$;}
            \State{$Y=\left[y^{(1)}, y^{(2)}, \cdots, y^{(N)}\right]$;}
            \State{$H_{T}=0,H_{T|Y}=0,S=[\;]$;}
            
            \Function {discrete\_entropy}{$L$}
                \State {Find the different elements in $L$, $v^{(1)},v^{(2)},\cdots,v^{(K)}$, and corresponding occurrence times, $o^{(1)},o^{(2)},\cdots,o^{(K)}$; }
                \State{Calculate the entropy by empirical distribution as
                \begin{equation}
                \label{eq: discrete_entropy_calculation}
                   H = - \sum_{i=1}^{K}  \frac{o^{(i)}}{len(L)} \log \frac{o^{(i)}}{len(L)};
                \end{equation}
                }\
                \State {\Return{$H$.}}
            \EndFunction
            \Statex{} % with no line-no
            
            \For{ $j=1 \rightarrow N$}
                \State{
                Discrete the continuous value as
                \begin{equation}
                    t^{(j)}_i = 
                    \lfloor 
                    \frac{t^{(j)}_i}{\lambda}
                    \rfloor;
                \end{equation}}
            \EndFor
            
            \State{$H_{T}$ = \Call{discrete\_entropy}{$T_i$};}
            \For{$m=1 \rightarrow M$}
                \State{$S[m]=$ find all elements whose class label is $m$ in $T_i$;}
                \State{$H=$\Call{discrete\_entropy}{$S[m]$};}
                \State{Calculate the conditional entropy as 
                    \begin{equation}
                       H_{T|Y} = H_{T|Y}+ \frac{len(S[m])}{N} \cdot H;
                    \end{equation}
                }
            \EndFor
            \State{\Return{$\hat{I}(T_i;X)=H_{T}$, $\hat{I}(T_i;Y)=H_{T}-H_{T|Y}$}.}

        \end{algorithmic}
    \end{algorithm}

\section{Experiments}
\label{sect: experiments}
\begin{figure*}[]
	\centering
    \subfloat[LeNet-5 on MNIST]{
		\centering
		\begin{minipage}[c]{1 \textwidth}
			\includegraphics[width=0.495 \textwidth]{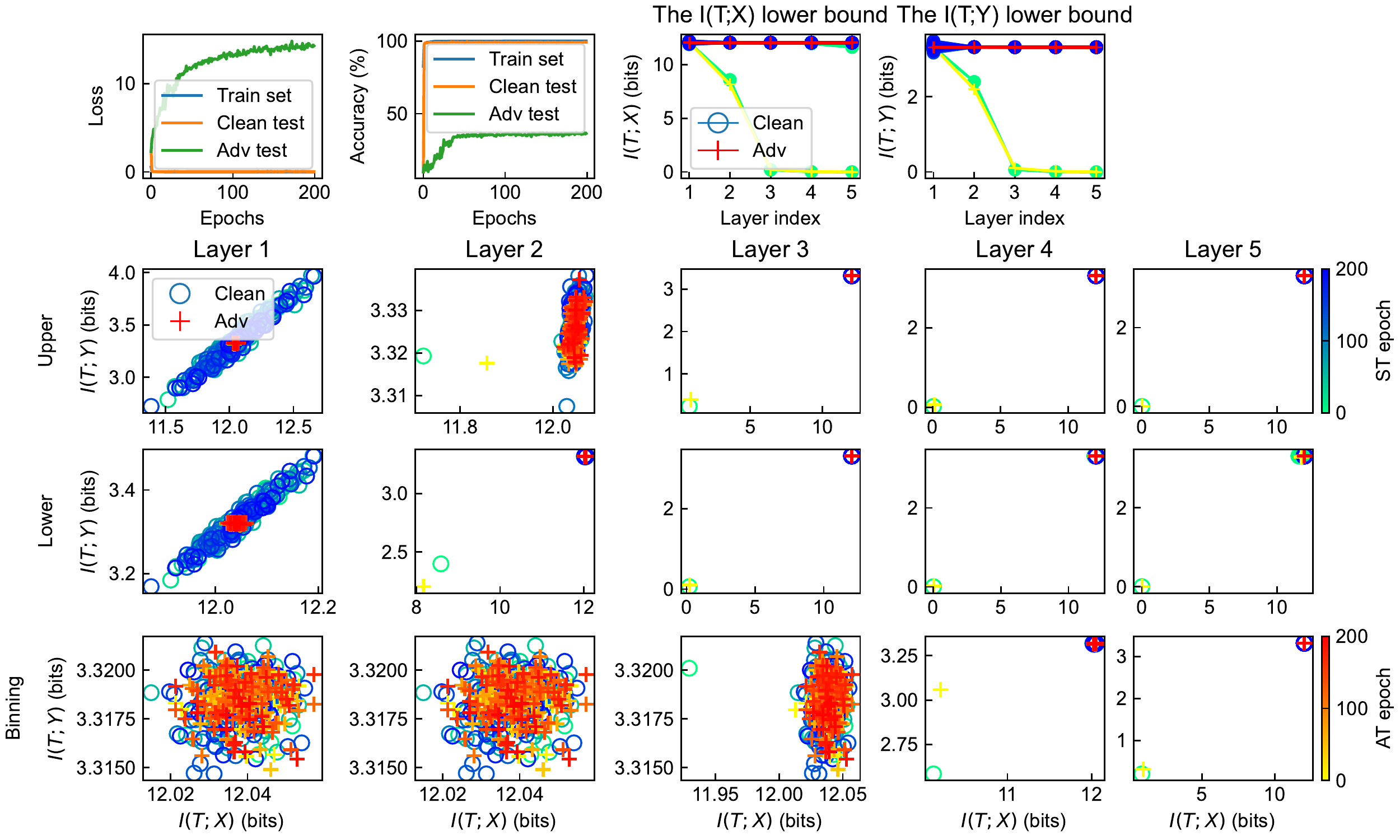}
			\includegraphics[width=0.495 \textwidth]{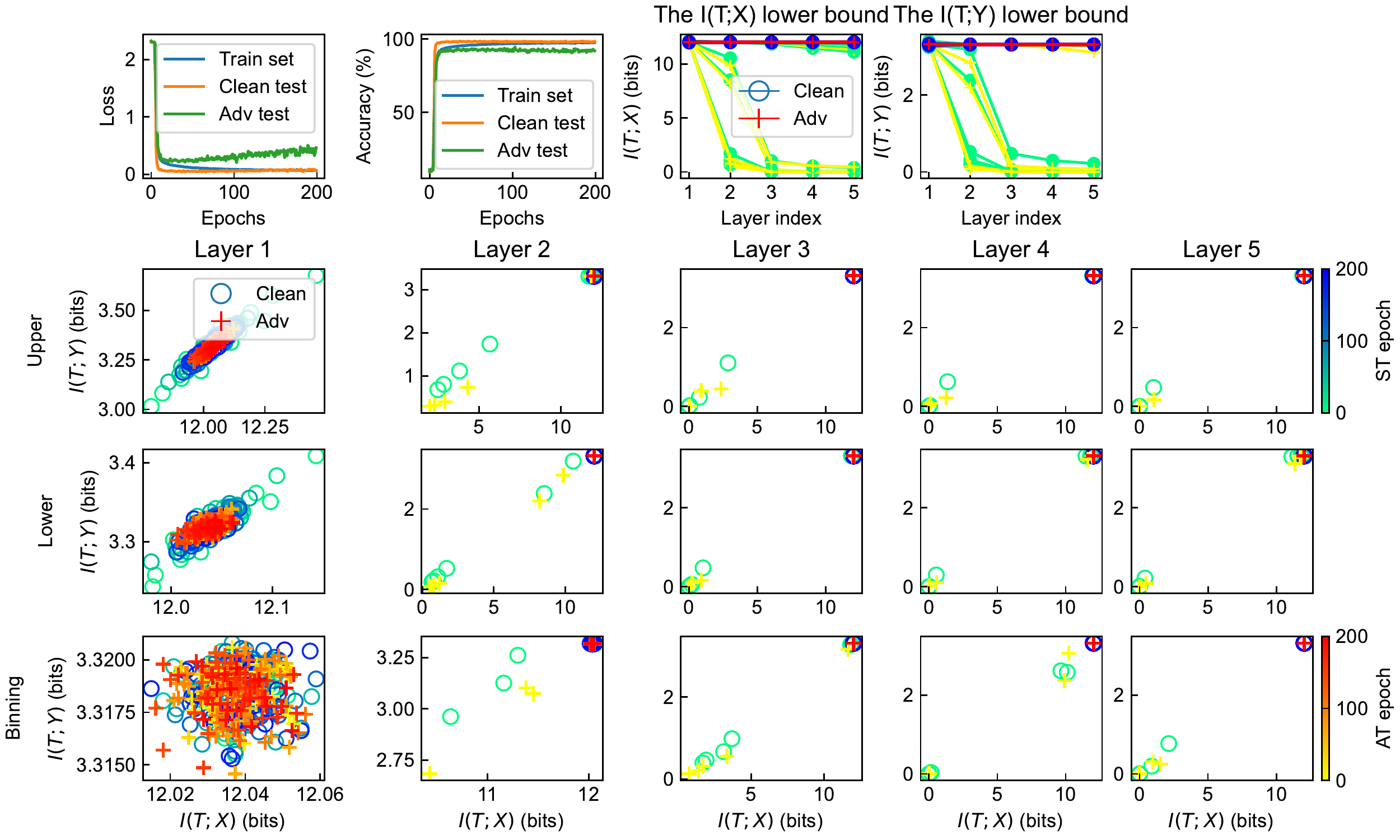}
		\end{minipage}
	}
	\\
    	\subfloat[WideResNet on CIFAR-10]{
		\centering
		\begin{minipage}[c]{1 \textwidth}
			\includegraphics[width=0.495 \textwidth]{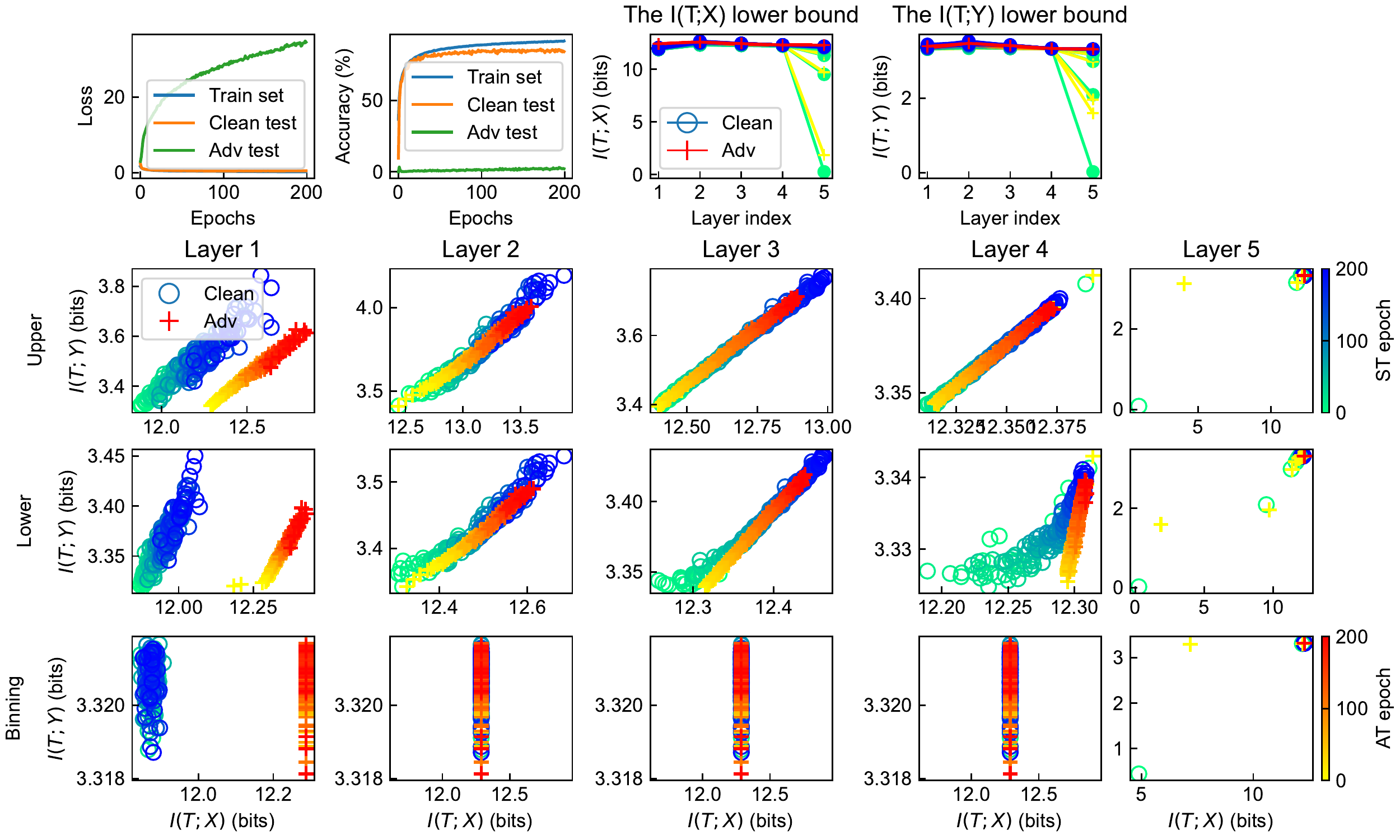}
			\includegraphics[width=0.495 \textwidth]{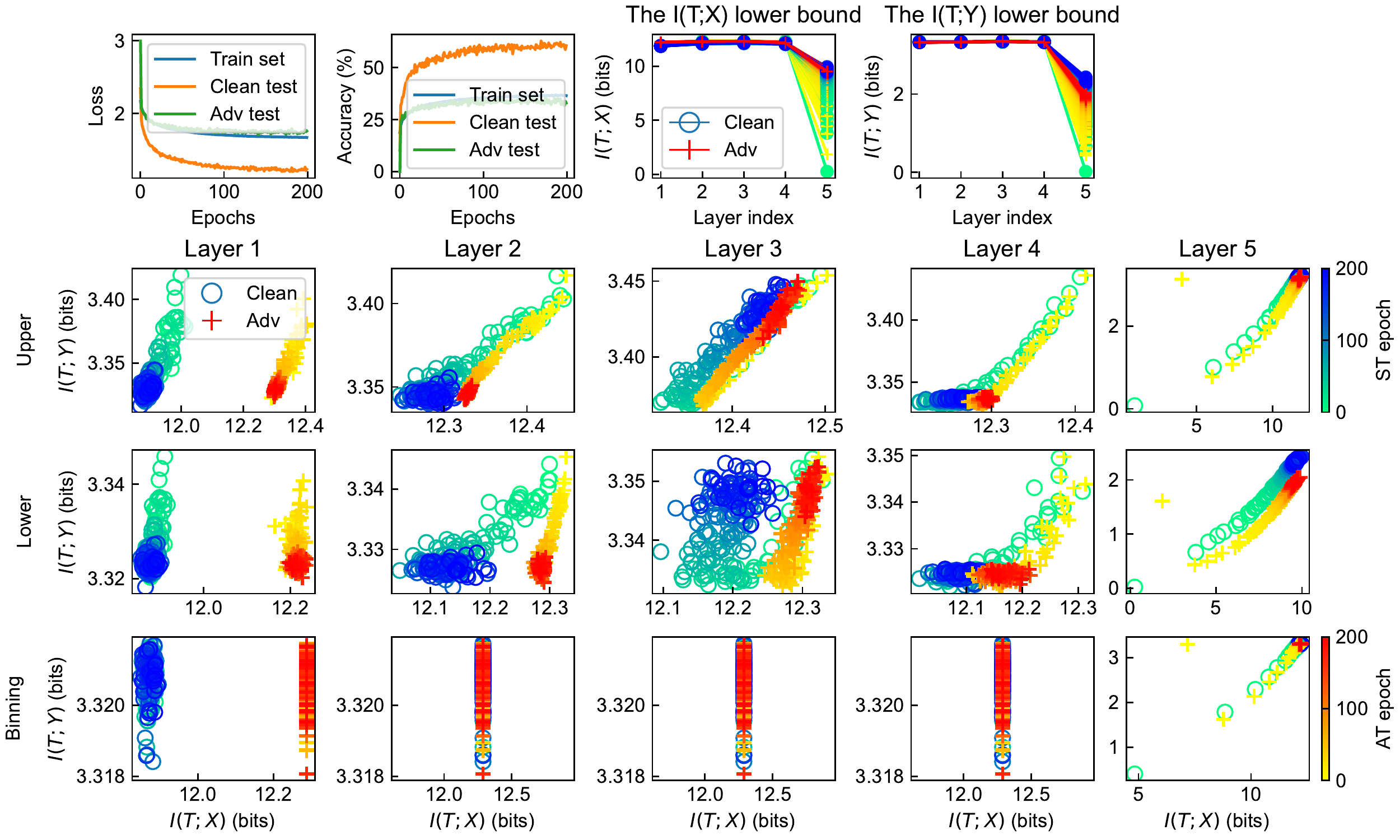}
		\end{minipage}
	}
	\\
		\subfloat[WideResNet on STL-10]{
		\centering
		\begin{minipage}[c]{1 \textwidth}
			\includegraphics[width=0.495 \textwidth]{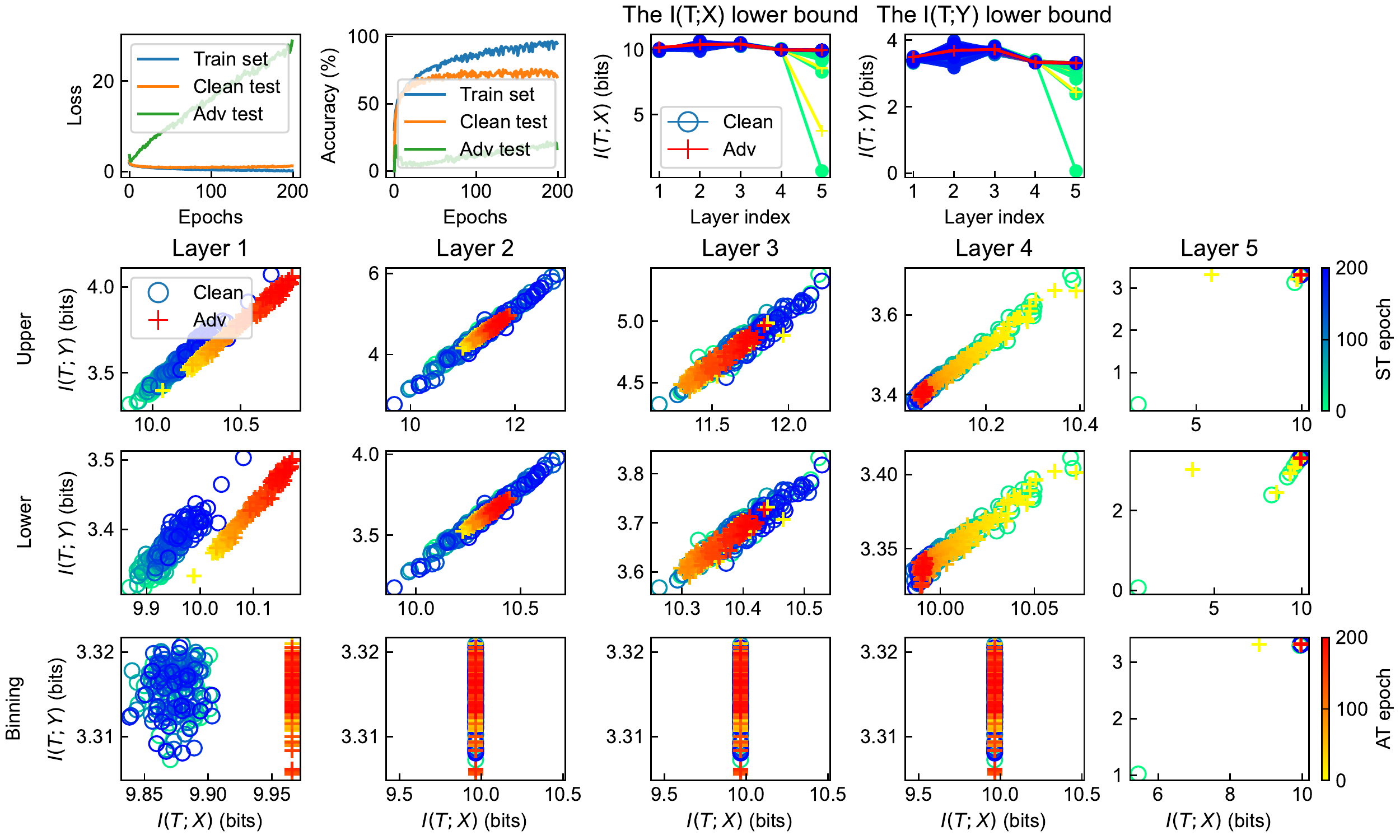}
			\includegraphics[width=0.495 \textwidth]{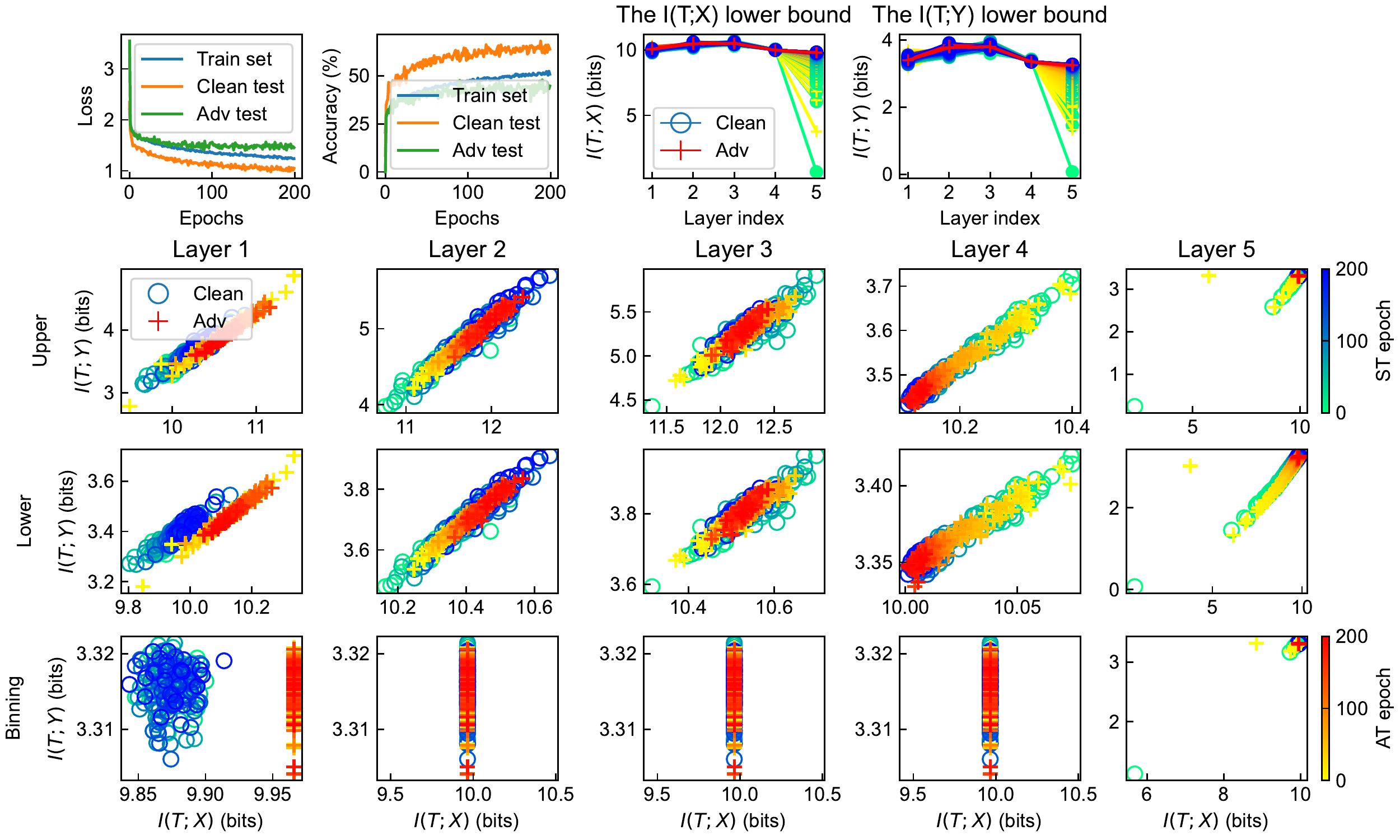}
		\end{minipage}
	}
	\caption{The information planes during normal training and adversarial training. The left column represents the result with normal training, and the right column represents the one with adversarial training. }
	\label{fig: experimental_results_1}
\end{figure*}

\begin{figure*}[]
	\subfloat[WideResNet on CIFAR-10]{
		\centering
		\begin{minipage}[c]{1 \textwidth}
			\includegraphics[width=1 \textwidth]{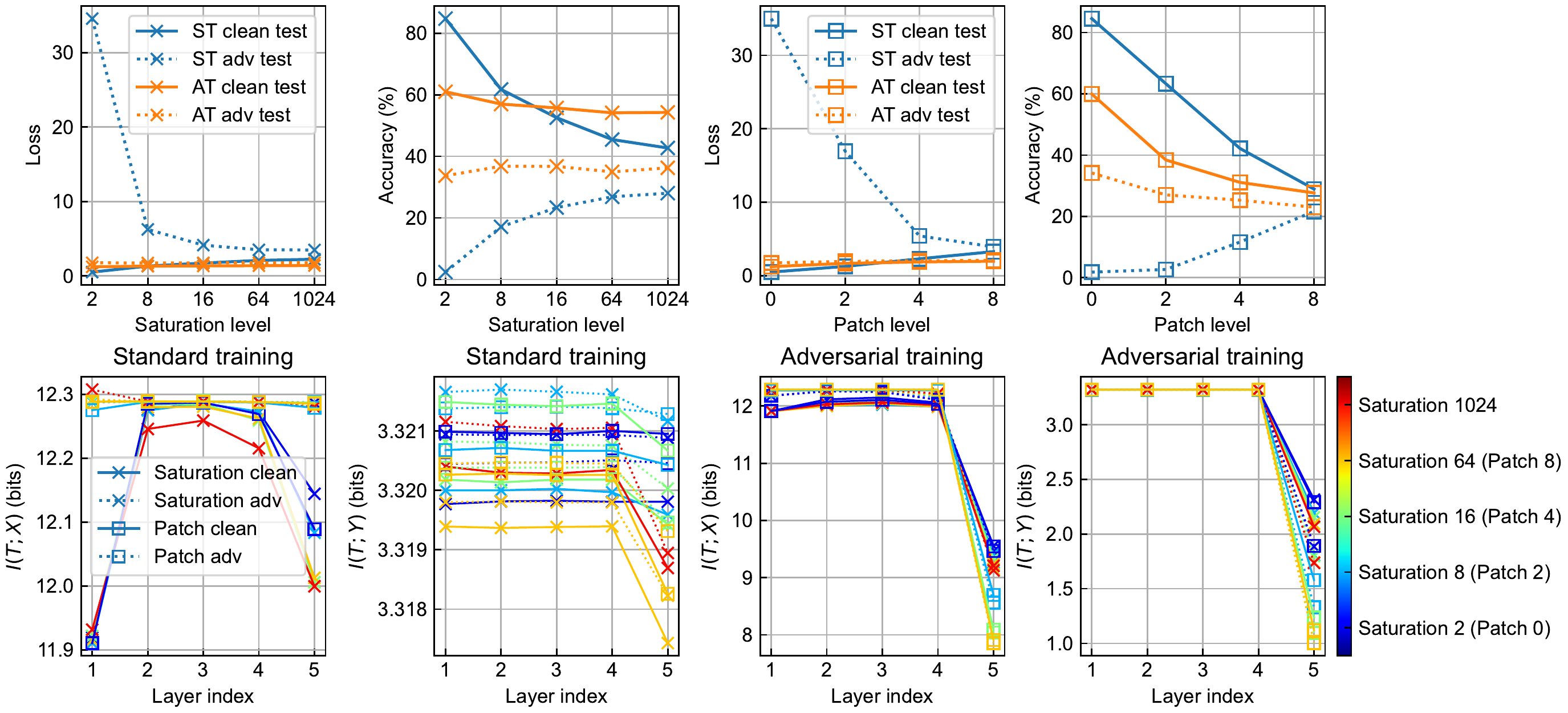}
		\end{minipage}
	}
	\\
		\subfloat[WideResNet on STL-10]{
		\centering
		\begin{minipage}[c]{1 \textwidth}
			\includegraphics[width=1 \textwidth]{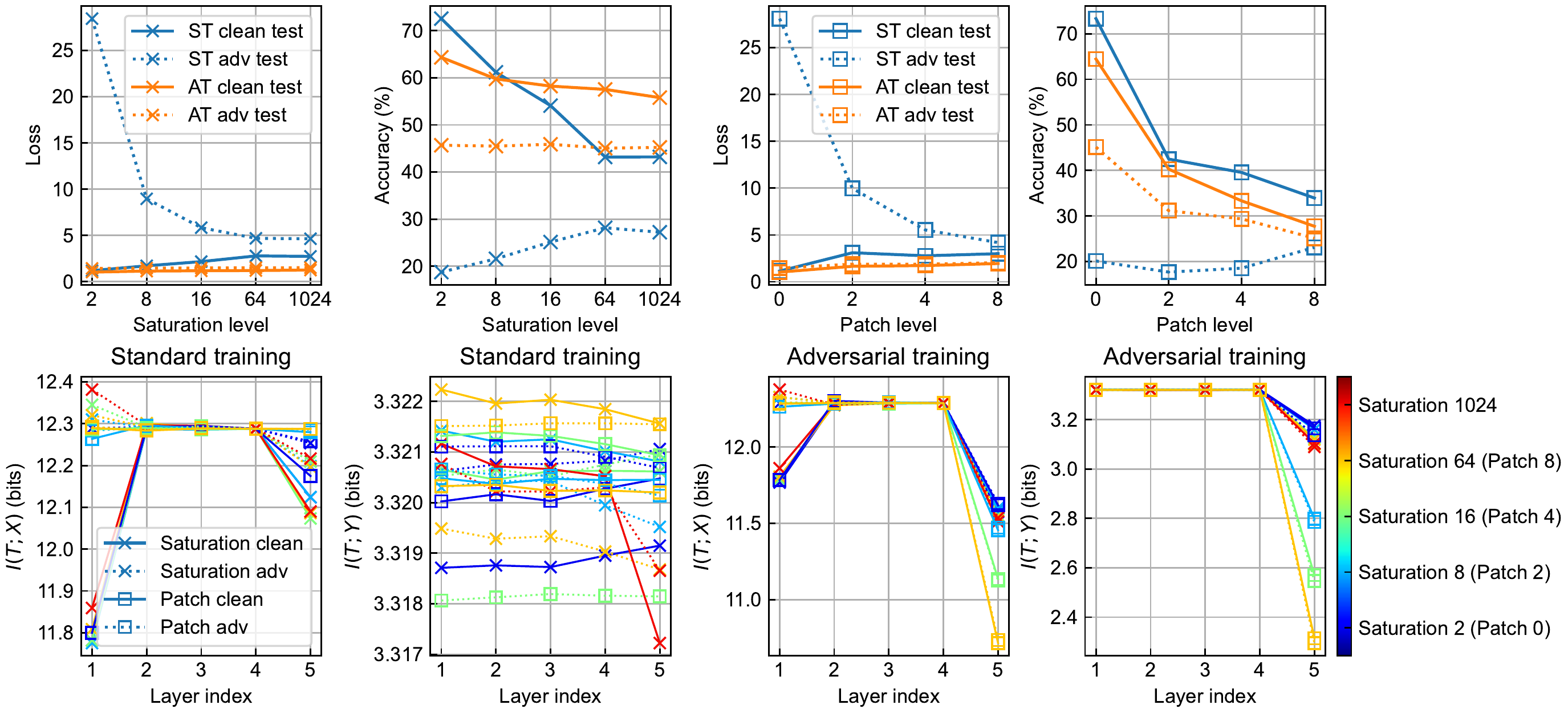}
		\end{minipage}
	}
	\caption{The plots of $I(T_i; X)$ and $I(T_i; Y)$ when the input suffers from different types of information distortion.}
	\label{fig: experimental_results_2}
\end{figure*}

% tables
\begin{table}[]
\caption{normal training and adversarial training parameters.}
\label{tab: experimental_setting}
\begin{tabular}{|l|ccc|}
\hline
Parameter     & \multicolumn{1}{c|}{MNIST Model} & \multicolumn{1}{c|}{CIFAR-10 Model} & STL-10 Model \\ \hline
Epochs        & \multicolumn{3}{c|}{200}                                                            \\ \hline
Optimizer     & \multicolumn{3}{c|}{SGD}                                                            \\ \hline
Batch Size    & \multicolumn{3}{c|}{128}                                                            \\ \hline
Learning Rate & \multicolumn{3}{c|}{0.1}                                                            \\ \hline
Momentum      & \multicolumn{3}{c|}{0.9}                                                            \\ \hline
Milestone     & \multicolumn{3}{c|}{20, 60}                                                         \\ \hline
Gamma         & \multicolumn{3}{c|}{0.5}                                                            \\ \hline
Epsilon       & \multicolumn{1}{c|}{45/255}      & \multicolumn{1}{c|}{8/255}         & 4/255       \\ \hline
Alpha         & \multicolumn{1}{c|}{8/255}       & \multicolumn{1}{c|}{2/255}         & 2/255       \\ \hline
Step          & \multicolumn{1}{c|}{7}           & \multicolumn{1}{c|}{7}             & 7        \\ \hline
\end{tabular}
\end{table}

% Please add the following required packages to your document preamble:
% \usepackage{multirow}
% Please add the following required packages to your document preamble:
% \usepackage{multirow}
\begin{table}[]
\caption{The details of Network architectures. $\star$ denotes the layer that will be probed to obtain the output.}
\label{table: network_architectutes}
\begin{tabular}{|c|c|c|}
\hline
LeNet-5 MNIST                                                                                                                                                                                               & WideResNet CIFAR-10                                                                                                                                                                                                                                                                                                                                                              & WideResNet STL-10                                                                                                                                                                                                                                                                                                                                                                 \\ \hline
\begin{tabular}[c]{@{}c@{}}Conv (5 $\times$ 5,  6) $\star$\\ Conv  (5 $\times$ 5,  16) $\star$\\ FC (400  $\times$ 120) $\star$\\ FC (120  $\times$ 84) $\star$\\ FC (84  $\times$ 10) $\star$\end{tabular} & \begin{tabular}[c]{@{}c@{}}Conv (3 $\times$ 3 $\times$16) $\star$\\ Conv  (3 $\times$ 3 $\times$16) $\star$\\ Conv  (3 $\times$ 3 $\times$16) \\ Conv  (3 $\times$ 3 $\times$32) $\star$\\ Conv  (3 $\times$ 3 $\times$32) \\ Conv  (3 $\times$ 3 $\times$64) $\star$\\ Conv  (3 $\times$ 3 $\times$64) \\ AvgPool  (8 $\times$ 8 ) \\ FC  (64 $\times$ 10) $\star$\end{tabular} & \begin{tabular}[c]{@{}c@{}}Conv (3 $\times$ 3 $\times$16) $\star$\\ Conv  (3 $\times$ 3 $\times$16) $\star$\\ Conv  (3 $\times$ 3 $\times$16) \\ Conv  (3 $\times$ 3 $\times$32) $\star$\\ Conv  (3 $\times$ 3 $\times$32) \\ Conv  (3 $\times$ 3 $\times$64) $\star$ \\ Conv  (3 $\times$ 3 $\times$64) \\ AvgPool  (8 $\times$ 8) \\ FC (576  $\times$ 10) $\star$\end{tabular} \\ \hline
\end{tabular}
\end{table}

\subsection{Empirical Setting}
\subsubsection{Datasets}
The train and test sets of MNIST \cite{LeNet-5_and_MNIST}, CIFAR-10 \cite{CIFAR-10}, and STL-10 \cite{STL-10} are used in this work. 
The MNIST's train and test sets, both from 10 classes, contain 60,000 and 10,000 $1 \times 28 \times 28$ images, respectively.
The CIFAR-10's train and test sets, both from 10 classes, contain 50,000 and 10,000 $3 \times 32 \times 32$ images, respectively.
The STL-10's train and test sets, both from 10 classes, contain 5,000 and 8,000 $3 \times 96 \times 96$ images, respectively.
\subsubsection{Models}
Two classic models are trained in this section: (a) the LeNet-5 \cite{LeNet-5_and_MNIST}, a simple CNN, achieves over 98\% top-1 accuracy on MNIST; (b) the WideResNet \cite{WideResNet} based on ResNet \cite{ResNet}, improvement with respect to performance and training speed by increasing network width,  achieves over 85\% top-1 accuracy on CIFAR-10 and STL-10.\
The architectures of the models are specified in Table \ref{table: network_architectutes}.
To reduce the amount of data, we do not obtain the output of all layers, but instead, select some important layers. In this work, for each model, the mutual information is estimated on 5 layers, as specified in Table \ref{table: network_architectutes}.
\subsubsection{Adversarial attack and Training} Adversarial attack and adversarial training are both based on the PGD method. But for different datasets and models, parameter settings for generating adversarial examples will be slightly different, the details as shown in Table \ref{tab: experimental_setting}.
\subsubsection{Saturation and Patch-Shuffling}
The saturation mainly changes the texture-based information of the image. For each pixel $v$ in an image, we change the saturation of $v$ as follows
\begin{equation}
\label{eq: saturation_operation}
v^{\prime}=\operatorname{sign}(2 v-1) \times |2 v-1|^{\frac{2}{p}} \times 0.5+0.5,
\end{equation}
where $v^{\prime}$ denotes the pixel with the saturation and $p$ is the saturation level. When $p=2$, the original image does not undergo saturation and the saturation adjustment range is $[2, 8, 16, 64, 1024]$. 
\par
The patch-shuffling mainly changes the shape-based information of the image. For an image, we first evenly split it into $k \times k$ patches, and randomly combine these $k \times k$ patches into a new image. Note that when $k=0$, it means that the original image will not be split. The size of $k$ is set to $[0, 2, 4, 8]$.

\subsection{Information Flows in Normal Training and Adversarial Training}
In this section, to explore the differences and similarities from the information perspective between NT-CNN and AT-CNN in their responses to the original and adversarial examples, we conduct experiments on three data sets: MNIST, CIFAR-10, and STL-10; and two models: LeNet-5 and WideResNet. During the normal training and adversarial training, for different examples (\textit{i,e}, original and adversarial examples), we used estimators mentioned in Section \ref{sect: mutual_info_estimators} to estimate the mutual information $I(T_i; X)$ between the $i$-th middle layer's output $T_i$ and the input $X$, and the mutual information $I(T_i; Y)$ between the $i$-th middle layer's output $T_i$ and the label $Y$ in each training epoch. 
When calculating mutual information, 5,000 examples are fed into the models to make the empirical distribution closer to the real distribution and achieve more accurate estimation results. 
At the same time, the classification accuracy and average cross-entropy loss of the model in the train set and test set are recorded during training. The experimental results are shown in Fig. \ref{fig: experimental_results_1}. It is worth noting that the last two columns of the first row of each subfigure overview the trends towards mutual information of each layer during training. Then the last three rows are information planes (the $x$-axis represents $I(T_i; X)$ and the $y$-axis represents $I (T_i; Y)$) of each layer, calculated by the three estimators, respectively. 
We obtain the following observations after analyzing the experimental results.
\begin{itemize}
    \item{With different training methods, over different datasets, on different CNN architectures, and estimating on different middle layers, there is a great diversity in the trends of the mutual information $I(T_i; X)$ and $I(T_i; Y)$. However, whether original or adversarial examples are the input,  the trends of the mutual information are nearly consistent when other settings are the same.
    For example, for WideResNet normally trained on CIFAR-10,  the trends of $I(T_i; X)$ and $I(T_i; Y)$ at all layers show a growth. When adversarially training on CIFAR-10, trends vary across layers.
    When WideResNet is trained on STL-10, the trends towards $I(T_i; X)$ and $I(T_i; Y)$ of some layers are increasing (\textit{e.g.}, the 1-st layer and 5-th layer) and ones of some layers are obscure. However, the trend on the original examples is almost in line with the one on adversarial examples. 
    % but the final value of mutual information for adversarial examples is weaker (mild). 
    }
    \item{The compression phase claimed in \cite{information_bottleneck_1} is almost not observed and the data processing inequality is not always clear.}
    \item{In contrast with NT-CNNs, the mutual information estimated in AT-CNNs, whether original or adversarial examples are as the input, shows a lagging advancement, and final results are at a lower level.}
\end{itemize}

% 在不用的训练模式下, 在不同的训练样本上, 在不同的模型上, 在不同的中间层进行互信息估计, 互信息的变化趋势展现出多样性. 但是, 对抗样本和普通样本的互信息估计结果趋势是一致的.
% which may be due to the small number of samples collected, only 1000 samples. This may be due to the small number of samples collected, only 1000 samples. 

% Models with high complexity seem to be very good at extracting information for simpler tasks, both for adversarial and normal training, and as accuracy increases, the value of mutual information quickly reaches its maximum. 

\subsection{Texture and Shape Distortion}
In this section, to explore the information extraction bias of NT-CNNs and AT-CNNs, we conducted experiments on two data sets, CIFAR-10 and STL-10, and one model, WideResNet. 
Concretely, for the model after 200 epochs of the normal and adversarial training, the examples (both original and adversarial examples) will suffer from texture or shape information distortion, then they are fed into the model. 
We calculate the lower bound of $I(T_i; X)$ and $I(T_i; Y)$ on the probed middle layers' output and simultaneously record the classification accuracy and average cross-entropy loss. 
The experimental results are shown in Fig. \ref{fig: experimental_results_2}. 
The first row of each subfigure displays the accuracy and average loss. The second row shows the mutual information of middle layers of models (both NT-CNNs and AT-CNNs) on examples suffering varying levels of texture-based and shape-based information distortions.
We obtain the following observations after analyzing the experimental results.
\begin{itemize}
\item{The patch-shuffling mainly changes the shape-based information contained in examples, but at the same time, the smaller patches are, the severely the texture-based information in the examples is distorted. 
The fact is the same for the saturation. As the saturation level increases, the texture-based information fades away and the shape information may also be wiped out.}

\item{The NT-CNNs are more sensitive to texture-based information and the AT-CNNs are more sensitive to shape-based information. 
The discrimination between the responses of the AT-CNNs to the patch-shuffling and saturation operation is obvious.
Whether the input is the original or adversarial example, as the patches become smaller,  $I(T_i; X)$ and $I(T_i; Y)$ at the last layer fall down.  Instead, the effect of the saturation is unapparent. Concretely, as the saturation level is enhanced, the mutual information shows slight decreases.}

\item{The NT-CNNs' responses to the patch-shuffling and saturation are more complicated. Specifically, higher saturation and smaller patches, for adversarial examples, may remove the texture-based information about other labels from the adversarial perturbations, and for original examples, may remove the texture-based information about the true label from themselves. 
For example, the experimental results of normally trained WideResNet on CIFAR-10 and STL-10 show that the higher the saturation, the smaller $I(T_i; X)$ at the last layer compared to the patch-shuffling.
At the same time, the mutual information curve of $I(T_i; Y)$ shows a rather complicated situation due to the two-sided effects of the saturation and patch-shuffling setting. }
% In addition,  $I(T_i; Y)$  of AT-CNNs is much lower than that of the NT-CNNs.}
\end{itemize}

\section{Discussion}
\label{sect: discussion}

\begin{figure*}[]
	\centering
	\includegraphics[width= \linewidth]{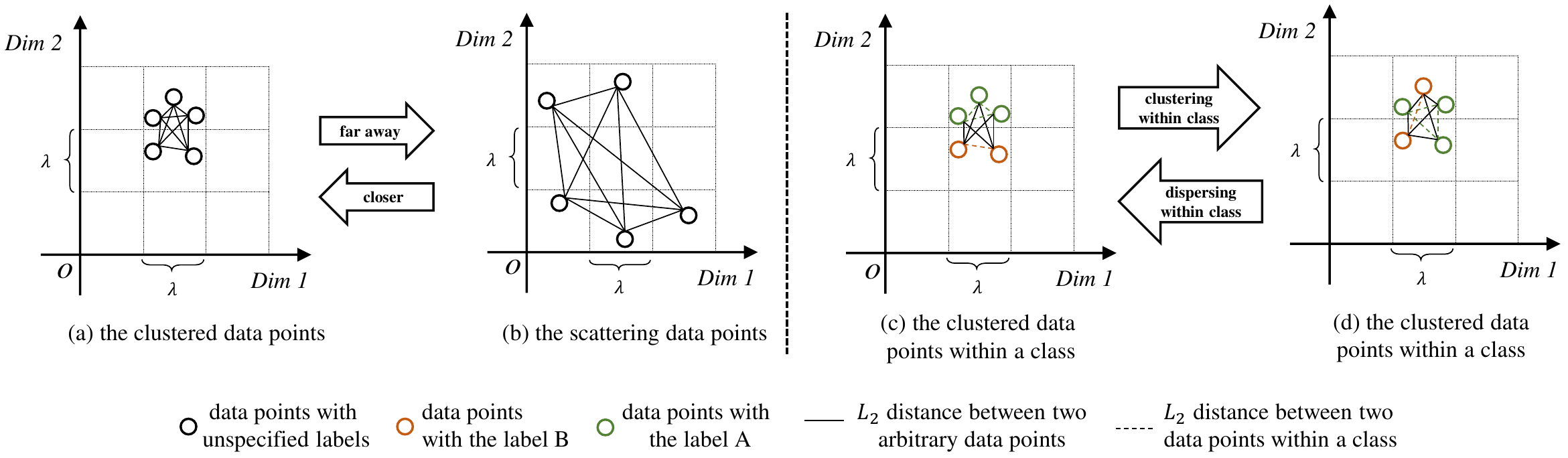}
    % 	Z 服从独立同分布的对角矩阵
	\caption{Illustration of the mutual information $I(T_i; X)$ and $I(T_i; Y)$estimated by the binning method in different distributions. $\lambda$ denotes the bin size. The circles represent the sampling result of the 2-dimensional random variable $T_i$, and each cell represents the 2-dimensional bin. 
	(a) The data points cluster tightly. The estimated mutual information $I\left(T_{i}; X\right)=H\left(T_{i}\right)=-\left(\frac{3}{5} \log \frac{3}{5}+\frac{2}{5} \log \frac{2}{5}\right) \approx 0.97$.
	(b) The data points scatter in the representation space. The estimated mutual information $I\left(T_{i} ; X\right)=-\left(\frac{1}{5} \log \frac{1}{5} \times 5\right) \approx 2.32$.
	(c) The data points within a class cluster tightly. The estimated mutual information $I\left(T_{i} ; Y\right)=H\left(T_{i}\right)-H\left(T_{i} \mid Y\right) =H\left(T_{i}\right)-\left(\frac{3}{5} \cdot 0+\frac{2}{5} \cdot 0\right) \approx 0.97$.
	(d) The data points within a class scatter. The estimated mutual information $I\left(T_{i} ; Y\right)=H\left(T_{i}\right)-H\left(T_{i} \mid Y\right) =H\left(T_{i}\right)-\left(\frac{3}{5} \cdot-\left(\frac{2}{3} \log \frac{2}{3}+\frac{1}{3} \log \frac{1}{3}\right)+\frac{2}{5} \cdot\left(\frac{1}{2} \log \frac{1}{2} \times 2\right)\right) =H\left(T_{i}\right)-\left(\frac{2}{5} \log \frac{3}{2}+\frac{1}{5} \log 3+\frac{2}{5} \log 2\right) \approx 0.02 $.
	}
	\label{fig: cluster_or_not}
\end{figure*}

Recall the previous formulas \eqref{eq: itx_by_kde_method} in the KDE method and \eqref{eq: discrete_entropy_calculation} in the binning method for estimating the mutual information. We simplify \eqref{eq: itx_by_kde_method} and have
\begin{equation}
\label{eq: simple_itx_by_kde_method}
\begin{aligned}
&\hat{I}(T_i;X) \\
&= 
-\log \frac{1}{N}
-\frac{1}{N} 
\sum_{j}
\log
\sum_{k} 
\exp \left(-\frac{1}{2} \frac{\left\|t^{(j)}_{i}-t^{(k)}_{i}\right\|_{2}^{2}}{\gamma \cdot \sigma^{2}}
\right).
\end{aligned}
\end{equation}
It is not difficult to find that $||t^{(j)}_{i}-t^{(k)}_{i}||_{2}^{2}$ is the key part of \eqref{eq: simple_itx_by_kde_method}. Namely, the Euclidean distance between $t^{(j)}_i$ and $t^{(k)}_i$ decides the value of the mutual information.
Meanwhile, when calculating $I(T_i; X)$ by the binning method the entropy of $X$ is calculated.
If the bin size is specified, the empirical distribution depicted by sampling data points, the foundation of the entropy, is closely related to the Euclidean distance between $t^{(j)}_i$ and $t^{(k)}_i$.
\par
Specifically, given $N$ examples $\{x^{(j)}\}_{j=1,\cdots,N}$ sampled from $P(X)$ and corresponding $N$ $i$-th middle layer's output $\{t^{(j)}_i\}_{j=1,\cdots,N}$, the mutual information $I(T_i;X)$ calculated by the KDE method is closely related to the Euclidean distance between $t^{(j)}_i$ and $t^{(k)}_i$. 
The farther two data points are from each other, the larger the mutual information $I(T_i; X)$ will be.
The fact is the same for the binning method, the empirical distribution will be closer to the uniform distribution as long as a farther distance between $t^{(j)}_i$ and $t^{(k)}_i$, and the mutual information calculated by the binning method will be larger, which is shown in Fig. \ref{fig: cluster_or_not}(a) and (b).
\par
$I(T_i; Y)$ is mostly similar to $I(T_i; X)$, but is not only related to the overall distribution of $T_i$, but also influenced by the distribution of $T_i$ within each class.
When the distribution of the overall $T_i$ is determined (\textit{i.e.}, $H(T_i)$ is known), for each class, the closer the distance between $t^{(j)}_i$ and $t^{(k)}_i$ within this class, the smaller the $H(T_i \mid Y)$,  the larger the $I(T_i;Y)$ estimated by KDE and binning methods, which is a little different from $I(T_i;X)$ and shown in Fig. \ref{fig: cluster_or_not}(c) and (d).
\par
According to the above analysis and Fig. \ref{fig: cluster_or_not}, the mutual information $I(T_i; X)$ and $I(T_i; Y)$ computed by the KDE and the binning methods can directly or indirectly portray the geometric properties of the $i$-th middle layer's output $T_i$.

\section{Conclusion}
\label{sect: conclusion}
In this work, we explored the middle layer's information plane of models during training with different training methods, on different types of examples, and on different CNN architectures. 
We also investigate the information curves while NT-CNNs and AT-CNNs receive the input that suffers from information distortion. 
We preliminarily study the mechanisms behind adversarial examples and adversarial training from the mutual information perspective. 
With different training methods, CNNs can be seen as information extractors with different information extraction biases.
The phenomena observed in previous experiments \cite{texture_bias_in_CNN_2019,texture_bias_in_CNN_2020,adversarial_examples_are_features,interpreting_AT-CNN} are also reflected from the mutual information perspective.
The KDE and binning estimators are not only a measure of the mutual information but also can depict the geometric properties of the middle layer's output based on the Euclidean distance and empirical distribution.
In the future, we will utilize more accurate mutual information estimators to obtain more revealing results. Besides, it is also significant to explore adversarial examples and DNNs' robustness from the geometric properties of the middle layer's output.
The resistance to adversarial examples in stochastic DNNs \cite{variational_information_bottleneck,nonlinear_information_bottleneck} also deserves our attention \cite{a_closer_look_at_VIB}. 

\appendices
\section{For discrete variables}
\label{appendix: A}
% 对于离散随机变量 $X$,以及离散变量 $T=f(X)$, 我们有 $H(T \mid X)=0$,  $I(T;X) = H(T)$. $I(T; X)$ 的展开式如下
For the sake of readability, we suppress the layer index $i$. For the discrete random variable $X$, and the discrete variable $T=f(X)$, we have $H(T \mid X)=0$, and $I(T;X) = H(T)$ as long as $H(T)$ is finite. The expression of $I(T; X)$ is as follows
\begin{equation}
\begin{aligned}
    I(T;X) &= H(T)-H(T \mid X) \\
         & = - \underset{t \in \mathcal{T}}{\sum} p(t) \log p(t) + \underset{x \in \mathcal{X}}{\sum} p(x) \underset{t \in \mathcal{T}}{\sum} p(t \mid x) \log p(t \mid x).
\end{aligned}
\end{equation}
% 因为$T$ 为 $X$ 的确定性函数且 $X=x$ 时,  变量 $T$ 的概率密度函数为
Because $T$ is a deterministic function of $X$, when $X=x$, the probability density function $p(t \mid x)$ is 
\begin{equation}
{p(t|x)} = 
\begin{cases}
1, &{t=f(x)},
\\ 
{0,}&{t \neq f(x)}.
\end{cases}
\end{equation}
% 并且我们定义 $0 \log 0 =0$, 则 $H(T \mid X)$ 为
We define $0 \log 0 =0$, so $H(T \mid X)$ is 
\begin{equation}
\begin{aligned}
    H(T \mid X)  &= - \underset{x \in \mathcal{X}}{\sum} p(x) 
    \Big (  p(f(x) \mid x) \log p(f(x) \mid x) \Big) \\
    &= - \underset{x \in \mathcal{X}}{\sum} p(x) 
    \Big(  1 \log 1 \Big)  = 0,
\end{aligned}
\end{equation}
and $I(T;X) = H(T)$.

\section{For continuous variables}
\label{appendix: B}
\begin{figure}[!h]
\centering
\includegraphics[width= 0.7 \linewidth]{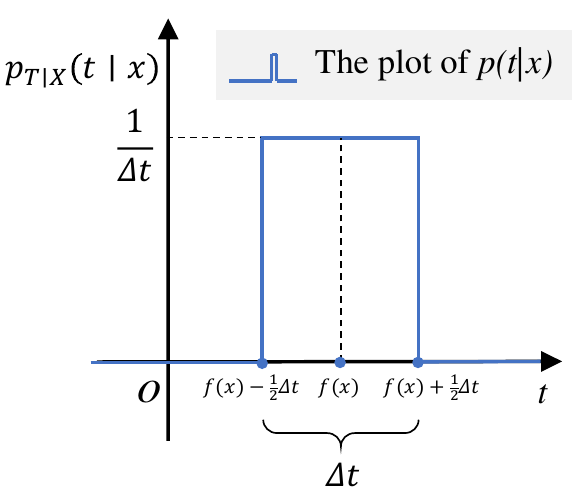}
\caption{The plot of probability density function $p(t \mid x)$.}
\label{fig: delta_function_like}
\end{figure}
% 当变量 X=x 时, 隐藏层变量T的分布函数 是delta 函数, 为了证明这一点, 我们先假设 隐藏层变量T的分布函数 如图 delta_function_like 所示, 当 \Delta 趋向于 0时, 我们可以得到 H(T|X)的计算结果趋向于 +无穷
For the continuous random variable $X$, and the continuous variable $T=f(X)$, we have $H(T \mid X)= - \infty$, and $I(T;X) = + \infty$ as long as $H(T)$ is finite. The expression of $I(T; X)$ is as follows
\begin{equation}
\begin{aligned}
    I(T;X) &= H(T)-H(T \mid X) \\
           & = - \int p(t) \log p(t) dt  \\ 
           & + \int p(x) dx  \int p(t \mid x) \log p(t \mid x) dt.
\end{aligned}
\end{equation}
The expression of $H(T \mid X)$ is as follows
\begin{equation}
    H(T \mid X) = - \int p(x) dx  \int p(t \mid x) \log p(t \mid x) dt.
    \label{eq: h(t|x)}
\end{equation}
As shown in Fig. \ref{fig: delta_function_like}, while $X=x$, we assume the distribution of $p(t \mid x)$ is as following
\begin{equation}
    p(t \mid x) = 
    \begin{cases}
        \frac{1}{\Delta t}, &{\lvert t-f(x) \rvert \leq \frac{1}{2} \Delta t},
        \\ 
         0, &{\lvert t-f(x) \rvert > \frac{1}{2} \Delta t},
    \end{cases}
    \label{eq: p(t|x)}
\end{equation}
and $\Delta t$ is a very small and positive value.
For the probability density function $p(x)$, we have 
\begin{equation}
    \int p(x) dx = 1, 
    \label{eq: integral_for_PDF}
\end{equation}
After plugging \eqref{eq: p(t|x)} and \eqref{eq: integral_for_PDF} into \eqref{eq: h(t|x)}, we have
\begin{equation}
\begin{aligned}
H(T \mid X) &= - \int p(x) dx  \int p(t \mid x) \log p(t \mid x) dt \\
       &=  - \lim_{\Delta t \rightarrow 0^+} \left( \Delta t \cdot \frac{1}{\Delta t} \cdot \log \frac{1}{\Delta t} \right) \\
        &= - \lim_{\Delta t \rightarrow 0^+} \left( \log \frac{1}{\Delta t} \right)
         = - \infty. \\
\end{aligned}
\end{equation}
This means that $H(T \mid X) \rightarrow -\infty$ and $I(T ; X) \rightarrow +\infty$ while $p(t \mid x)$ approaches the delta function.

 % argument is your BibTeX string definitions and bibliography database(s)
% \bibliography{IEEEabrv,../bib/paper}
%

% \bibliographystyle{IEEEtran}
\bibliographystyle{ieeetr}
\bibliography{IEEEabrv,references} 
% \begin{thebibliography}{1}
% \bibliographystyle{IEEEtran}

\begin{IEEEbiography}[{\includegraphics[width=1in,height=1.25in,clip,keepaspectratio]{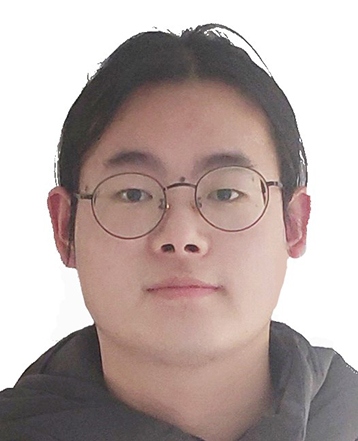}}]{Jiebao Zhang} received the B.E. degree in information management \& information system from Anhui University of Technology, Anhui, China, in 2020. He is currently pursuing the master’s degree in computer science with the School of Information Science and Engineering, Yunnan University, Kunming 650500, China.
His research interests include deep learning,  AI security, and explainable AI.
\end{IEEEbiography}

\begin{IEEEbiography}[{\includegraphics[width=1in,height=1.25in,clip,keepaspectratio]{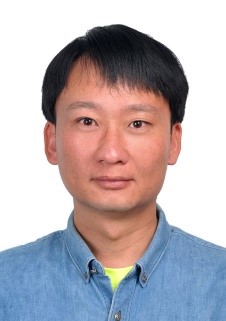}}]{Wenhua Qian} received the M.S. degree from Yunnan University, Kunming, China, in 2005, and the Ph.D. degree from Yunnan University, Kunming, China, in 2010. He is currently a professor at Computer Science and Engineering Department of Yunnan University, Kunming China. From 2017 to 2021, he was a Postdoctoral Research Fellow with the Department of Automation, Southeast University, Nanjing, China. He has authored or co-authored over 60 papers in refereed international journals. His current research interests include computer vision, image processing, and cultural computing. 
\end{IEEEbiography}

\begin{IEEEbiography}[{\includegraphics[width=1in,height=1.25in,clip,keepaspectratio]{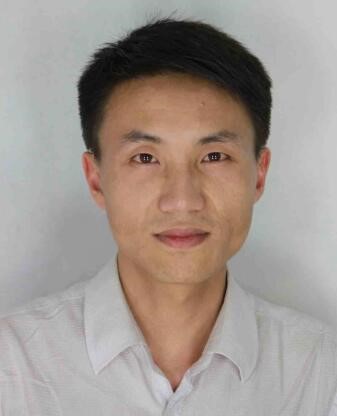}}]{Rencan Nie} received the B.E. degree in electronic information science and technology and the M.S. and Ph.D. degrees in communication and information systems from Yunnan University, Kunming, China, in 2004, 2007, and 2013, respectively. Since 2016, he has been a Post-Doctoral Research Fellow with the Department of Automation, Southeast University, Nanjing, China. He is currently a Professor at the Department of Communication Engineering, Yunnan University. He has authored or co-authored over 60 papers in refereed international journals. His current research interests include neural networks and image fusion.
\end{IEEEbiography}

\begin{IEEEbiography}[{\includegraphics[width=1in,height=1.25in,clip,keepaspectratio]{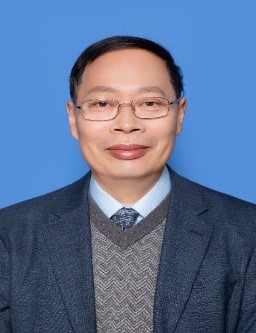}}]{Jinde Cao (Fellow, IEEE)} received the B.S. degree from Anhui Normal University, Wuhu, China, the M.S. degree from Yunnan University, Kunming, China, and the Ph.D. degree from Sichuan University, Chengdu, China, all in mathematics/applied mathematics, in 1986, 1989, and 1998, respectively. He is an Endowed Chair Professor, the Dean of the School of Mathematics, the Director of the Jiangsu Provincial Key Laboratory of Networked Collective Intelligence of China, and the Director of the Research Center for Complex Systems and Network Sciences at Southeast University. Prof. Cao was a recipient of the National Innovation Award of China, the Gold medal of the Russian Academy of Natural Sciences, the Obada Prize, and the Highly Cited Researcher Award in Engineering, Computer Science, and Mathematics by Thomson Reuters/Clarivate Analytics. 
He is elected as a member of the Academy of Europe, a foreign member of the Russian Academy of Sciences, a foreign member of the Russian Academy of Engineering, a member of the European Academy of Sciences and Arts, a foreign fellow of the Pakistan Academy of Sciences, a fellow of African Academy of Sciences, a foreign member of the Lithuanian Academy of Sciences, and an IASCYS academician.
\end{IEEEbiography}

\begin{IEEEbiography}[{\includegraphics[width=1in,height=1.25in,clip,keepaspectratio]{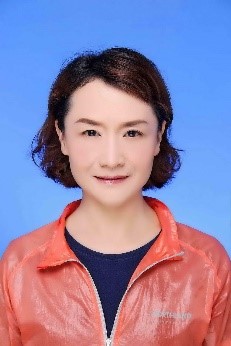}}]{Dan Xu} received the M.Sc. and Ph.D. degrees in computer science and technology from Zhejiang University, Hangzhou, China, in 1993 and 1999, respectively. She is currently a Professor at the School of Information Science and Engineering, Yunnan University, Kunming, China. She has authored or co-authored over 100 papers in refereed international journals. Her research interests include different aspects of image-based modeling and rendering, image processing and understanding, computer vision.
\end{IEEEbiography}

\end{document}